\pdfoutput=1

\documentclass[11pt]{article}

\usepackage[]{acl}

\usepackage{times}
\usepackage{latexsym}
\usepackage{multirow}
\usepackage{amsmath, amsfonts, amssymb}
\usepackage{tabularx}
\usepackage{booktabs}
\usepackage{graphicx}
\usepackage{colortbl}
\usepackage{makecell}
\usepackage{multirow}
\usepackage{multicol}
\usepackage{booktabs}
\usepackage{adjustbox}
\usepackage{graphicx}
\usepackage{arydshln}
\usepackage{listings}
\usepackage{tablefootnote}
\usepackage{CJKutf8}
\usepackage{enumitem}

\usepackage[T1]{fontenc}

\usepackage[utf8]{inputenc}

\usepackage{microtype}

\usepackage{inconsolata}

\title{Role Prompting Guided Domain Adaptation with General Capability Preserve for Large Language Models}

\author{Rui Wang$^{\heartsuit}$,  Fei Mi$^{3*} $, Yi Chen$^{\heartsuit}$, Boyang Xue$^{1, 2}$, \\
\textbf{Hongru Wang$^{1, 2}$, Qi Zhu$^{3}$, Kam-Fai Wong$^{1, 2}$, Ruifeng Xu$^{\heartsuit}$\thanks{\ \ Corresponding Author.}}\\
  $^{\heartsuit}$Harbin Insitute of Technology, Shenzhen, China\\
  $^{1}$MoE Key Laboratory of High Confidence Software Technologies, China\\
  $^{2}$The Chinese University of Hong Kong $^{3}$Huawei Noah's Ark Lab \\
  {\tt ruiwangnlp@outlook.com, mifei2@huawei.com, xuruifeng@hit.edu.cn}
 }

\begin{document}
\maketitle
\begin{abstract}

The growing interest in Large Language Models (LLMs) for specialized 
applications has revealed a significant challenge: when tailored to specific domains, LLMs tend to experience catastrophic forgetting, compromising their general capabilities and leading to a suboptimal user experience. Additionally, crafting a versatile model for multiple domains simultaneously often results in a decline in overall performance due to confusion between domains.
In response to these issues, we present the \textbf{R}ol\textbf{E} Prompting \textbf{G}uided Multi-Domain \textbf{A}daptation (REGA) strategy. This novel approach effectively manages multi-domain LLM adaptation through three key components:
\textbf{1) Self-Distillation} constructs and replays general-domain exemplars to alleviate catastrophic forgetting. 
\textbf{2) Role Prompting} assigns a central prompt to the general domain and a unique role prompt to each specific domain to minimize inter-domain confusion during training.
\textbf{3) Role Integration} reuses and integrates a small portion of domain-specific data to the general-domain data, which are trained under the guidance of the central prompt. 
The central prompt is used for a streamlined inference process, removing the necessity to switch prompts for different domains.
Empirical results demonstrate that {REGA} effectively alleviates catastrophic forgetting and inter-domain confusion. This leads to improved domain-specific performance compared to standard fine-tuned models, while still preserving robust general capabilities.

\end{abstract}
\begin{CJK*}{UTF8}{gbsn}
\section{Introduction}

Large Language Models (LLMs) \cite{LLAMA, LLAMA2, GPT3, InstructGPT} have revolutionized the field of Natural Language Processing, demonstrating exceptional general capabilities, such as instruction-following \cite{InstructGPT, flan} and complex reasoning \cite{cot}. 
However, general-purpose LLMs might fall short in some specific areas requiring professional knowledge, due to the lack of exposure to data in relevant domains. 
Hence, there has emerged an increasing number of studies in developing domain-specific models by injecting domain knowledge into LLMs in some domains, e.g., medicine \cite{pmc-llama,huatuo}, law \cite{ChatLaw}, and finance \cite{bloomberggpt,xuanyuan}.

\begin{figure}[t]
\centering
\includegraphics[width=0.5\textwidth]{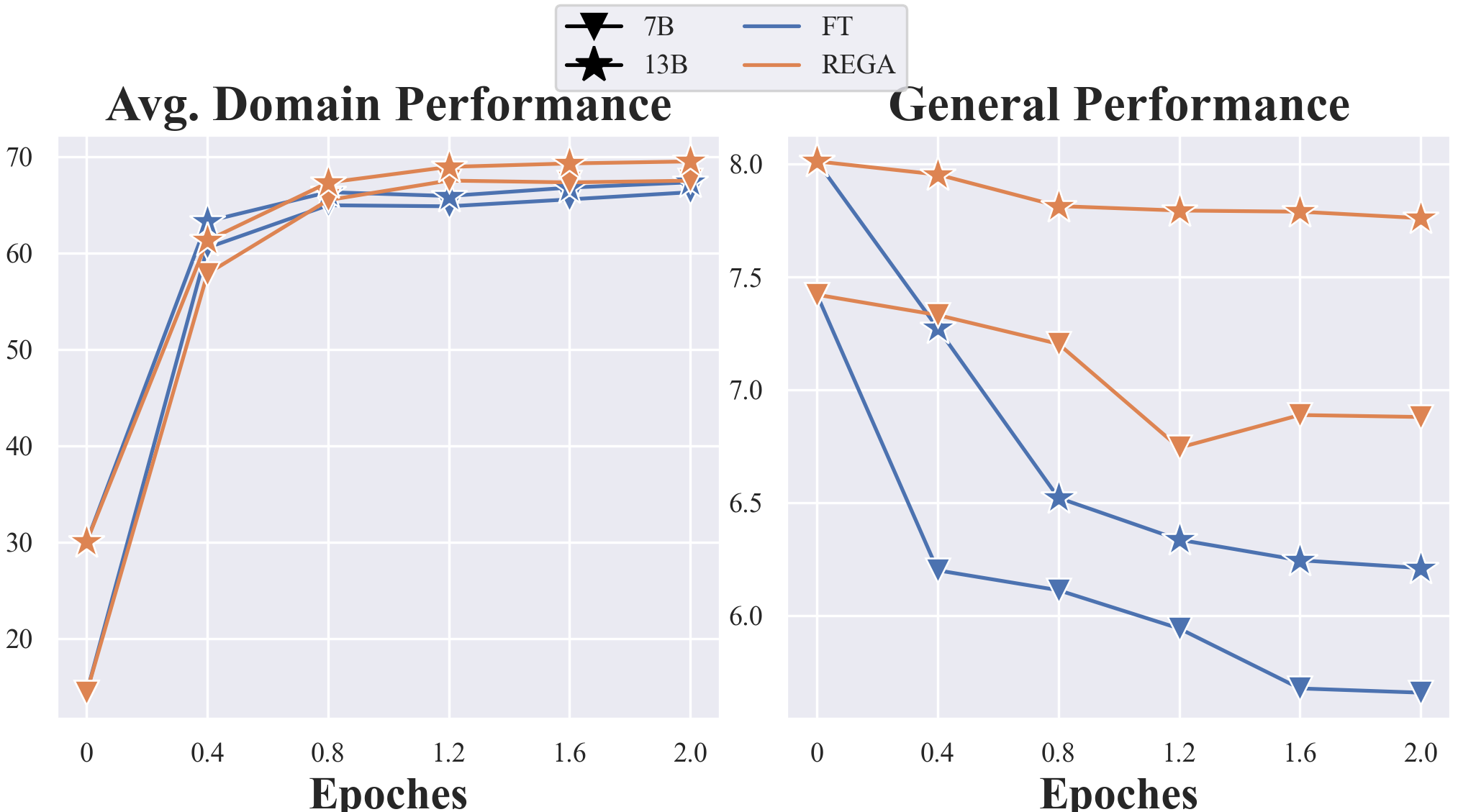}
\caption{Performance Comparison of {BELLE} with varied sizes tuned by  {Standard Finetuning (FT)} and {REGA}. 
The models tuned by {FT} suffer from a severe drop in general performance as the training epoch increases. Whereas, the counterparts tuned by {REGA} are better at preserving general capacities while achieving comparable domain-specific performance.
}
\label{fig:FT_trend}
\vspace{-2mm}
\end{figure}

Nevertheless, adapting LLMs to specific areas risks triggering \textit{catastrophic forgetting} \cite{specializing_cpst,LSTM_fg, cf1,xuanyuan,MMLM_fg}.
As shown in Figure~\ref{fig:FT_trend}, the enhancement of specialized abilities comes at the cost of the generic ability to follow diverse instructions. 
This dilemma underscores the need for effective solutions that uphold the balance between domain-specific mastery and general applicability. 
Besides, directly adapting a single LLM to multiple domains simultaneously through standard finetuning 
may cause \textit{inter-domain confusion} \cite{multi_domain_training}, which negatively affects the model performance in each specific domain.

\begin{figure*}[t]
\centering
\includegraphics[width=1\textwidth]{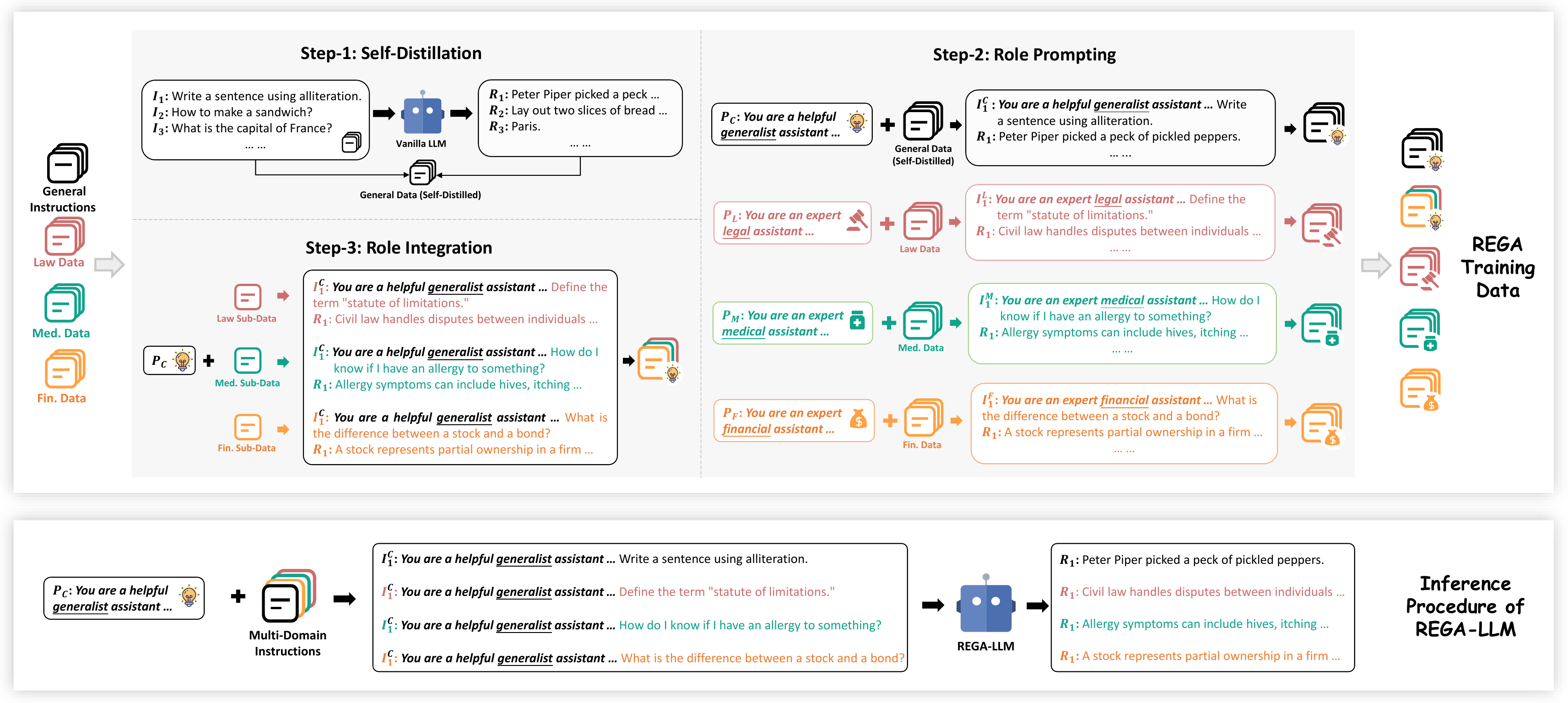}
\caption{
Overview of REGA. 
\textbf{For training}, REGA organizes the training data by: 
(1) \textit{Self-Distillation}: The vanilla LLM generates exemplars according to a set of general-domain instructions to preserve generic abilities.
(2) \textit{Role Prompting}: The LLM is assigned a unique role through role prompts, which are concatenated with samples in corresponding domains. $P_C$ is the central prompt indicating the generalist role for the general domain, while $P_L$, $P_M$, and $P_F$ are the expert role prompts for law, medicine, and finance domains.
(3) \textit{Role Integration}: 
A fraction of data from each specialized domain is mixed with the general-domain data, all guided by the central prompt, which integrates various expert roles into the generalist role. 
\textbf{For inference}, the central prompt effectively guides the LLM tuned on REGA training data to respond to multi-domain instructions, without the need for role prompt selection.
}
\label{fig:method}
\vspace{-2mm}
\end{figure*}

To this end, we propose the \textbf{R}ol\textbf{E} Prompt \textbf{G}uided Multi-Domain \textbf{A}daptation (\textbf{REGA}) strategy. 
As shown in Figure~\ref{fig:method}, given the instruction-following pairs from multiple target domains and our collected general-domain instructions, {REGA} reconstructs the training data for robust multi-domain adaptation through three key steps.

\noindent \textbf{(1) 
Self-Distillation} leverages the LLM itself to generate responses to the pre-collected diverse general-domain instructions before domain adaptation.
The distilled instruction-following exemplars will be rehearsed during training to retain the generic abilities of the LLM, without the need to access the original, often private pre-training data. 

\noindent \textbf{(2) Role Prompting} assigns the LLM with a unique expert role when adapting to distinct professional domains, and a generalist role by default when tackling general-domain data. This is done by concatenating a role prompt to the beginning of corresponding domain-specific or general-domain instructions. 
The role prompts act as system guidance to inform the LLM of clear domain boundaries during training, thus alleviating inter-domain confusion.

\noindent \textbf{(3) Role Integration} samples a small portion of data from each target domain and reuses them for training, all under the guidance of the central prompt.
By guiding model training on the common domain-specific data, the different domain-sensitive expert roles are transferred and integrated into the generalist role of the central prompt.

\noindent During the inference stage, we directly use the central prompt to guide the model to handle instructions from various domains smoothly, alleviating the burden of role prompt engineering.

We conduct extensive experiments by adapting several LLMs in both Chinese and English datasets spanning three domains, including medicine, law, and finance.
The experiment results exhibit that LLMs trained with REGA surpass other baselines in domain performance by a large margin while having a significant generic performance advantage.
Furthermore, our detailed analysis underscores the effectiveness of each component of REGA. 
We reveal strong evidence that {Self-Distillation} is a reliable method for preventing the loss of general capabilities (\S~\ref{sec:self-d}).
Additionally, Role Prompting is critical in reducing inter-domain confusion (\S~\ref{sec:role-p}).
Lastly, Role Integration proves to be vital for the successful incorporation of knowledge from specific domain roles into a unified central role (\S~\ref{sec:role-i}), which is essential for the model's adaptability.

\section{Related Work}

\paragraph{Catastrophic Forgetting}
It has been observed that domain-specific tuning of LLMs can lead to catastrophic forgetting \cite{cf1, cf2}, where an LLM loses its ability to perform previously learned tasks effectively. 
This suggests a balance must be struck between domain specialization and general proficiency.
To mitigate catastrophic forgetting, particularly in the context of continual learning, researchers have explored three kinds of strategies. 
\textit{Exemplar replay} involves preserving and revisiting key training examples to maintain model performance \cite{exemplar-1, exemplar-2}. 
\textit{Regularization} methods introduce regulation functions in addition to the loss function to constrain the learning process \cite{cf1, regu1}. 
\textit{Architectural methods} adjust the model's structure by adding parameters specific to new tasks or domains \cite{arch1}. 
Our task setting is to train an LLM that can competently handle multiple domains concurrently, with minimal impairment to its generalist capabilities, differentiating from continual learning where the model is exposed to tasks sequentially, striving to prevent significant forgetting of earlier tasks \cite{arch1}.

\paragraph{Inter-domain Confusion}
Furthermore, training a single LLM for multiple domains risks triggering \textit{inter-domain confusion} where the LLM may not perform as well in each domain due to the blending of domain-specific knowledge \cite{multi_domain_training, multidomain2}. 
Therefore, some studies have been directed toward identifying commonalities across domains to maintain model performance while preserving the unique characteristics of each domain \cite{multi_domain_training, multidomain2}.
In this paper, we propose to utilize Role Prompting to alleviate inter-domain confusion.

\paragraph{Role Prompting}
Previous works found that role prompting can significantly improve the performance of LLMs.
For example, Character.AI \footnote{https://beta.character.ai/} proposes a dialogue agent mimicking diversified figures, which can bring enriched user experience.
Moreover, \citet{role-eval} found LLMs can effectively evaluate summarization results with diversified role prompts from varied perspectives.
\citet{role-prompt-reasoning} found role prompting can also boost the complex reasoning abilities of LLMs.
Inspired by these findings, we propose to utilize role prompting to help LLMs distinguish samples among domains and assign domain-specific abilities to each role.
Our experiments demonstrate that role prompting can effectively alleviate inter-domain confusion.

\section{Method}
\vspace{-1mm}
\subsection{Preliminaries}

Consider that there is a large corpus whose domain distribution is known, which is $\mathcal{D} = \{\mathcal{D}_1, \mathcal{D}_2, ..., \mathcal{D}_n\}$, where each $\mathcal{D}_i$ encompasses several sub-datasets about the $i^{th}$ domain.
$D_i$ consists of instruction-response pairs, which means $(x_i,y_i) \in \mathcal{D}_i$. $x_i$ and $y_i$ represent the instruction and response respectively.

Our goal is to utilize $\mathcal{D}$ to train a language model $\theta$ to obtain $\theta'$ 
which has strong performance across $n$ domains simultaneously without considerably compromising its general performance capability.

\subsection{The REGA Tuning Strategy} \label{sec:REGA}
As shown in Figure~\ref{fig:method}, REGA is a framework for organizing training datasets from multiple domains to obtain a final training corpus, which can improve the domain performance of LLMs without considerably compromising its general performance capability.

\subsubsection{Self-Distillation}

To alleviate catastrophic forgetting in the general domain, a straightforward and effective method is selecting exemplars in the training data and replaying \cite{exemplar-1, exemplar-2} them to LLMs besides the domain-specific data.
However, the original training data of many LLMs are often proprietary and not open-sourced, so we try to partially replace it by devising the {Self-Distillation}.
Specifically, we first collect a set of high-quality instructions $\mathcal{I}=\{(x_g,)\}$ from the general domain and let the LLM $\theta$ generate responses $y_g$ for each $x_g$ (as shown in the Step-1 part of Figure~\ref{fig:method}). 
This generated dataset $\mathcal{I}=\{(x_g, y_g)\}$, henceforth referred to as $D_g$, 
which is preserved as exemplars in the general domain and will be replayed in the following training process to restore the model's generic knowledge distribution.
Now our training corpus can be denoted as $\mathcal{D}^+ = \{\mathcal{D}_g, \mathcal{D}_1, \mathcal{D}_2, ..., \mathcal{D}_n\}$

\subsubsection{Role Prompting}
Although the self-distilled $D_g$ can alleviate the catastrophic forgetting, directly training $\theta$ on $D^+$ will degrade its performance on each one \cite{multi_domain_training, multidomain2} due to confusion among domains.
To alleviate the inter-domain confusion, we introduce the {Role Prompting} to help LLMs distinguish among domains by assigning role prompts for data from each domain (as shown in the Step-2 part of Figure~\ref{fig:method}). 
In particular, the general domain is assigned a central prompt $p_c$, and each of $n$ domains is assigned a unique role-prompt, forming a role-prompt set $P=\{p_c, p_{1}, p_{2},..., p_{n}\}$. 
Then each instruction-responses pair $(x,y)$ is prefixed with its corresponding domain-specific role prompt, 
which means the current training dataset is
$\mathcal{D}_r^+=\{ (p_c \oplus x_g,y_g)| (x_g,y_g) \in \mathcal{D}_g\} \bigcup \{(p_{i} \oplus x_i,y_i) | (x_i,y_i) \in \bigcup_{i=1}^{n} \mathcal{D}_i \}$.

\subsubsection{Role Intergration}\label{sec:data-share}
The {Role Prompting} can segregate domain-specific data during the training process 
but it also makes it crucial to determine which role prompt to use based on the domain of the input.
To obviate the need for role prompt selection during inference, we design the {Role Intergration} that enables the central-prompt $p_c$ to acquire the specialized abilities associated with each domain's role-prompt $p_i$.
The key to this strategy is the reinforcement of the versatility of the central prompt, allowing the LLMs to process prompts from all domains using $p_c$.
Concretely, a fraction of data is randomly selected from each domain's dataset $D_i$, denoted as $D_i'$, is combined with the general domain data $D_g$ and prefixed with the central prompt $p_c$. 
The composite data collection is thus structured as $\mathcal{T}_r^s=\{(p_c \oplus x_g, y_g)|(x_g,y_g) \in D_g \cup (\bigcup_{i=1}^{n}D_i') , D_i'\subset D_i \} \bigcup \{(p_{i} \oplus x_i,y_i) |(x_i,y_i) \in\bigcup_{i=1}^{n} D_i \}$.

\subsubsection{Training Corpus of REGA}
The final training corpus that REGA builds upon $\mathcal{D}^+$ is $\mathcal{T}_r^s$.
Having trained the LLM $\theta$ on $\mathcal{T}_r^s$, we obtain the $\theta '$.
Besides, we introduce the mixing ratio $r$, quantifying the ratio of each selected subset $D_i'$ to its full domain dataset $D_i$. 
The mixing ratio is defined as $r = |D_i'|/|D_i|$.
This metric facilitates the calibration of domain exposure during the training process.

\subsection{The REGA Inference Procedure}
At the inference stage, we only need the central prompt to guide LLMs in the generation.
For the given input $x_u$, the prediction process is represented as $y_u = {\theta '}(p_c \oplus x_u)$.
This process bypasses the need for selecting different role prompts for each domain, thereby streamlining model deployment and ensuring consistency in responses across varied domains.

\begin{table*}[ht]
\centering
\small
\begin{tabular}{c|c|ccc|cc|cccccc}
\toprule
\multirow{2}{*}{Model} & General & \multicolumn{3}{c}{Medicine} & \multicolumn{2}{c}{Law} & \multicolumn{6}{c}{Finance} \\
\cmidrule(lr){2-2}\cmidrule(lr){3-5}\cmidrule(lr){6-7}\cmidrule(lr){8-13}

  & CGev & QQ & TC & MQA & LQA & LS & FNA & FQA & FNL & FRE & FFE &FSP\\ 
\midrule

\rowcolor[rgb]{0.93,0.93,0.93}\multicolumn{13}{c}{\textit{BELLE-7B}} \\
0-shot  & 7.42 & 8.80 & 1.60 & 12.32 & 16.88 & 9.85 & 49.54& 24.98 &41.73 &0.00 & 2.99& 18.13 \\

FT  & 5.41 & 83.10 & 75.94 & 36.84 &\textbf{58.05} &44.82 &61.07 &72.83 & 93.47& 50.75& 67.39& 85.41\\
FTSD  & 6.26 & \textbf{85.21} & 74.44& 32.64 & 57.04& 44.29& 59.81&75.10 &91.31 &47.76 & \textbf{69.57} &85.44\\
{REGA}$^c$  & \textbf{6.87} & 82.39 & \textbf{76.69} &\textbf{37.95} &57.65 & \textbf{46.90}& \textbf{61.68}&\textbf{78.18} &\textbf{95.65} &\textbf{55.22} &68.48 & \textbf{85.68}\\

\rowcolor[rgb]{0.93,0.93,0.93}\multicolumn{13}{c}{\textit{BELLE-13B}} \\
0-shot  & 8.01 & 33.10 & 3.01 & 41.72 & 56.47 & 35.15 & 41.53& 23.85 &41.30 &7.46 & 33.70& 13.20 \\

FT  & 6.21 &84.51 & \textbf{81.96} & 35.63 & 58.25 & 45.51 & 61.87 & 77.30 & 91.30 & {61.20} & 68.11 & 86.08 
\\
FTSD  & 6.92& 84.37 & 78.95 & 36.47 & \textbf{58.43} & 44.58 & 61.67 & 74.69 & \textbf{93.48} & 59.29 & 72.83 & 85.76  \\
{REGA}$^c$   & \textbf{7.75} & \textbf{85.33} & 79.22 & \textbf{37.67} & 58.11 & \textbf{47.52} & \textbf{62.27} & \textbf{77.79} & 92.33& \textbf{62.37}& \textbf{73.71} & \textbf{88.06}\\
\midrule
{Metrics} &- & Acc. & u.F1 & u.F1 & u.F1 & u.F1 & u.F1 & u.F1 & Acc. & Acc. & Acc. & u.F1\\

\bottomrule
\end{tabular}
\caption{We present the performance of BELLE in different experimental conditions, with the top scores highlighted in \textbf{bold}. 
The superscript $^c$ indicates that the model's assessment was conducted using the central prompt $p_c$.
Acc. or u.F1 means that the evaluation metric of this dataset is Accuracy or Uni-gram-F1 respectively. The mixing ratio of REGA is 0.1.}
\label{tab:Chinese_performance}
\end{table*}

\begin{table}[t]
\centering
\small
\begin{tabular}{c|c|p{0.77cm} p{0.77cm}|c|c}
\toprule
\multirow{2}{*}{Model} & \multicolumn{1}{c}{General} & \multicolumn{2}{c}{Medicine} & \multicolumn{1}{c}{Law} & \multicolumn{1}{c}{Finance} \\
\cmidrule(lr){2-2}\cmidrule(lr){3-4}\cmidrule(lr){5-5}\cmidrule(lr){6-6}
 & MTB & PMQA & MMQA & CQA & FQA \\ 
\midrule
\rowcolor[rgb]{0.93,0.93,0.93}\multicolumn{6}{c}{\textit{Vicuna-7B}} \\
0-shot & 6.23 & 42.68 & 31.28 & 18.60 & 24.02  \\

FT & 4.57& 52.17 & 42.07 & 66.80 & 32.17\\
FTSD & 5.68& 60.87 & \textbf{42.27} & 67.20 & 39.12\\
{REGA}$^c$   & \textbf{6.11} & \textbf{65.21} & {41.41} & \textbf{68.80} & \textbf{45.24}\\
\midrule
Metrics & - &Acc. & Acc. & Acc. & u.F1\\

\bottomrule
\end{tabular}
\caption{The performance of Vicuna-7B is detailed below, with the highest scores emphasized in \textbf{bold}. Acc. or u.F1 means that the evaluation metric of this dataset is Accuracy or Uni-gram-F1 respectively. The mixing ratio of REGA is 0.1.}
\label{tab:english_performance}
\end{table}

\section{Experiment}

\subsection{Datasets} 

In this section, we introduce the domain datasets we utilized and the instruction set for \textbf{Self-Distillation}.
We perform the experiments on three domains, medicine, law, and finance.
We choose datasets carefully to contain both language understanding and generation tasks for more comprehensive evaluation of LLMs. 
The statistics and detailed metrics of datasets are shown in Appendix~\ref{app:data_statis}.

\paragraph{English Datasets} We encompass four English datasets across the medical, legal, and financial domains, including {PubMedQA} \cite{PubMedQA}, {MedMCQA} \cite{MedMCQA}, {casehold\_QA} \cite{casehold}, and {FinBertQA}\footnote{https://sites.google.com/view/fiqa}.

\paragraph{Chinese Datasets} For the Chinese portion of our study, we have sourced 11 datasets from three different sectors. 
For medical, we include {cMedQQ} ({QQ}) for paraphrase identification, {cMedTC} ({TC}) for sentence classification \cite{zhang-etal-2022-cblue} and {cMedQA} ({MQA}) for question answering \cite{cmedQA}; in the legal domain, we have {LawQA} ({LQA}) for question answering\footnote{https://github.com/pengxiao-song/LaWGPT/tree/main} and {LawSum} ({LS}) for document summarization\footnote{http://cail.cipsc.org.cn}; and for finance, datasets such as {FNA}, {FQA}, {FNL}, {FRE}, {FFE}, and {FSP}, which cover a range of tasks from sentiment analysis to entity relation classification, are adopted from \citet{ZhFin-Lu2023a}.

\paragraph{General Instruction Datasets}
In light of existing research underscoring the importance of data quality and diversity while training LLMs \cite{textbook, self-inst}, 
we try to build high-quality and diversified instruction datasets to better preserve the models' generic capabilities in the constructed $D_g$ after {Self-Distillation}.
For the Chinese models, we have randomly extracted 50K instruction samples from both the Chinese-Alpaca\footnote{https://github.com/ymcui/Chinese-LLaMA-Alpaca} and MOSS \cite{moss} projects, resulting in a combined total of 100K samples.
In the case of the English models, we have likewise randomly chosen 50K instruction samples from each of the WizardLM \cite{wizardlm} and Alpaca\footnote{https://github.com/tatsu-lab/stanford\_alpaca} projects, amounting to a total of 100K samples.

Then these instructions are fed into the \texttt{BELLE} and \texttt{Vicuna} to obtain distilled exemplar set $D_g$. 
In the decoding process, we set the temperature to 0.7 and top-p to 0.95 for response generation.

\subsection{Role Prompt Setting}
We design role prompts for medicine, law, and finance domains respectively.
However, we use the central prompt $p_c$ in line with the original instruction-tuning process of the model rather than a fresh one. 
For instance, take the prompt used during Vicuna's instruction-tuning: "\textit{A chat between a curious user and an artificial intelligence assistant. The assistant is designed to be helpful, detailed, and polite in responding to user queries.}" 
This same prompt is employed as the central prompt $p_c$ in REGA to create our training dataset for Vicuna. 
Our goal of using the same $p_c$ as the one in the instruction-following process is to preserve the foundational knowledge the model originally had.

\subsection{Baselines}\label{sec:baselines}
\paragraph{Zero-Shot}
We evaluate the \texttt{BELLE} and \texttt{Vicuna} on the domain and general test set with greedy decoding in a zero-shot setting.

\paragraph{Standard Finetuning (FT)}
We finetune the LLM $\theta$ on domain-specific datasets spanning three distinct domains, respectively represented by $D_m$, $D_l$, and $D_f$ and the training corpus denoted as $T_{ft} = \{D_m \cup D_l \cup D_f\}$.
This finetuning process results in a refined model $\theta_{ft}$. 
For a given user input $x_u$, the inference stage of $\theta_{ft}$ is expressed as $y_u = {\theta_{ft}}(x_u)$.

\paragraph{Standard Finetuning with Self-Distillation (FTSD)} 
We combine {FT} and {Self-Distillation} to diagnose catastrophic forgetting while training with {FT} and explore the effects of {Self-Distillation}.
The {FTSD} approach integrates the self-distilled instruction-response dataset $\mathcal{D}_g$ into the fine-tuning corpus $\mathcal{D}$, resulting in the $\mathcal{T}_{ftsd} = \{\mathcal{D}_g \cup \mathcal{D}_m \cup \mathcal{D}_l \cup \mathcal{D}_f\}$. 
After training $\theta$ on $T_{ftsd}$, we obtain $\theta_{ftsd}$.
For a given user input $x_u$, the inference stage of $\theta_{ftsd}$ is denoted as $y_u = {\theta_{ftsd}}(x_u)$.

\paragraph{Standard Finetuning with Role Prompting (FTRP)}
We combine {FT} and {Role Prompting} to explore the existence of inter-domain confusion and the effects of {Role Prompting}.
We use the same training corpus $T_{ft}$ as {FT} in this setting but assign the role prompts to the instructions of each domain.
Assume we have role prompts $p_m$, $p_l$ and $p_f$ for medicine, law, and finance domains, the training corpus can be denoted as ${T}_{ftrp}=\{(p_{i} \oplus x_i,y_i) |(x_i,y_i) \in\bigcup_{i \in \{m,f,l\}} D_i \}$.
For a given user input $x_u$, we need to choose different role prompts according to the domain of $x_u$, while inferring with the obtained model $\theta_{ftap}$.
For example, if the $x_u$ is from the medical role prompt, the inference procedure is $y_u = {\theta_{ftrp}}(p_m \oplus x_u)$.

\begin{figure}[t]
\centering
\includegraphics[width=0.45\textwidth]{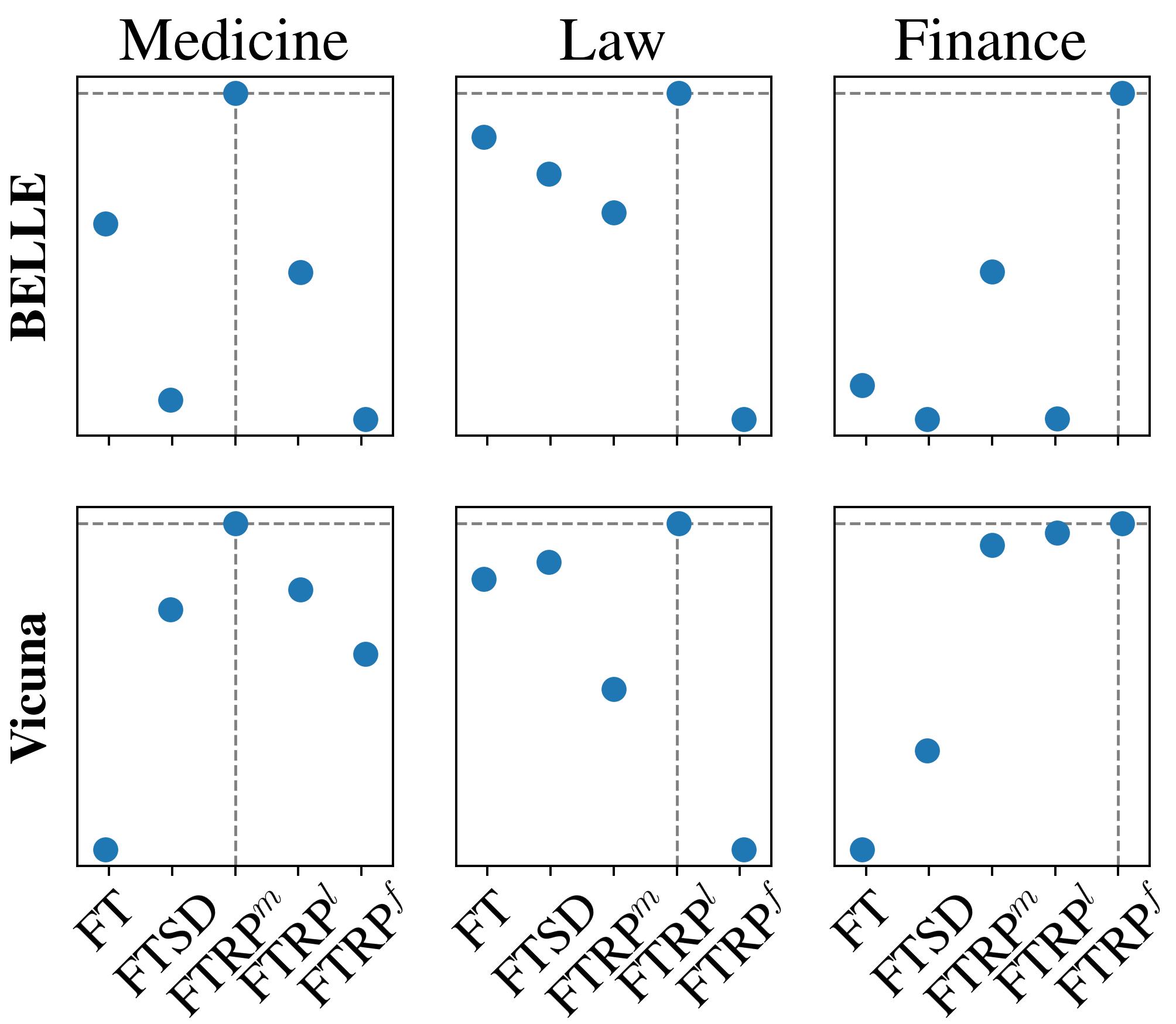}
\caption{Performance of BELLE and Vicuna tuned by {FTRP}. 
{FTRP}$^x$ indicate models are tested by the role prompts $p_x$ of the $x$ domain. 
}
\label{fig:FTRP_COMP}
\vspace{-2mm}
\end{figure}

\paragraph{REGA}
The {REGA} training and inference procedure are already described in Section~\ref{sec:REGA}.

For the existence of role prompts of {REGA} and {FTRP}, we also denote their inference process as {REGA}$^x$ and {FTRP}$^x$, which means that the model trained with {REGA} or {FTRP} are using the role prompt $p_x$ of the domain $x$.

\subsection{Models} For our experiments with Chinese datasets, we have selected models from the BELLE series \footnote{https://github.com/LianjiaTech/BELLE}, specifically \texttt{BELLE-7B-2M} and \texttt{BELLE-13B-2M}. 
These models are iterations of \texttt{LLaMA-7B} and \texttt{LLaMA-13B} respectively \cite{LLAMA}. 
They have been further fine-tuned in a supervised manner on a Chinese dataset containing 2 million instruction-response pairs.
Regarding English datasets, our choice of the base model is \texttt{Vicuna-1.5-7B} \footnote{https://github.com/lm-sys/FastChat}, which has been fine-tuned from \texttt{LLaMA2-7B} \cite{LLAMA2}.
We train these models in the LoRA \cite{LoRA} manner. 
The $r$ and $\alpha$ of LoRA are 16 and 32 respectively.
For all of the methods, batch size is set to 16, and the maximum number of epochs is set to 2.
We test performance on the checkpoint obtained after the second epoch.

\subsection{Evaluation}

For domain performance, we evaluate the models on the corresponding test datasets using automatic metrics, including accuracy and uni-gram-F1 (also illustrated in Table~\ref{tab:Chinese_performance} and Table~\ref{tab:english_performance}).

As for the general performance, we evaluate the English models on MT-Bench (MTB) \cite{mt-bench} and the prompt format follows the exact setting of MT-Bench.
Each response is evaluated by a numerical score ranging from 0 to 10.
For the Chinese models, we collect an evaluation collection, \textbf{CGev}, consisting of 650 samples, to test model abilities across coding, reasoning, question answering, classification, and conversation tasks. 
The distribution of the tasks in \textbf{CGev} and the prompt format are shown in Appendix~\ref{app:data_statis}.
We evaluate \texttt{BELLE} series models on \textbf{CGev} by asking GPT-4 to give a numerical score
(from 0 to 10) for the single response.
All of the model's general performances are automatically evaluated by GPT-4-0613 with greedy decoding to reduce randomness.

\begin{figure*}[t]
\centering
\includegraphics[width=1\textwidth]{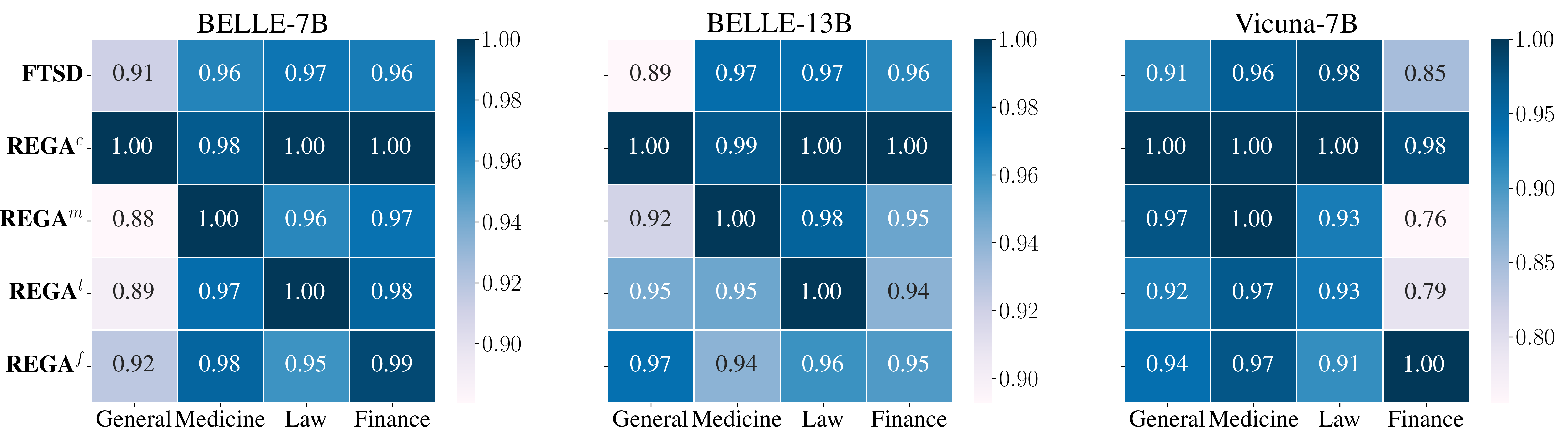}
\caption{We present the normalized performance metrics for the \texttt{BELLE} and \texttt{Vicuna-7B}, which are fine-tuned using the \textbf{FTSD} and \textbf{REGA}. The notation \textbf{REGA}$^x$ indicates that the model's inference is performed using the role prompt $p_x$.  The normalization process involves dividing each score by the maximum score within the same column. The mixing ratio of REGA is 0.1.
}
\label{fig:rega_role}
\end{figure*}


\subsection{Performance Analysis}\label{sec:diagnose}
To clearly illustrate the experiment results, 
we present the results of {Zero-shot} ({0-shot}), {FT}, {FTSD}, and {REGA} in Table~\ref{tab:Chinese_performance} and Table~\ref{tab:english_performance}.
The performance of {FTRP}, {FT} and {FTSD} are in Figure~\ref{fig:FTRP_COMP}.
Several interesting observations can be noted.

\paragraph{Diagnosing Catastrophic Forgetting} 
While {FT} can consistently improve domain performance, it tends to compromise the model's overall proficiency. 
As shown in Table~\ref{tab:Chinese_performance} and Table~\ref{tab:english_performance}, 
the general performance of these models decreases across languages and model sizes. 
In particular, the \texttt{BELLE-7B} model sees its score decrease from 7.42 to 5.41. 
Similarly, the \texttt{BELLE-13B} model's score declines from 8.01 to 6.21 after {FT}.

\paragraph{Diagnosing Inter-domain Confusion} 
To investigate the existence of inter-domain confusion, we fine-tune \texttt{BELLE} using only medical datasets. 
The outcomes, depicted in Table~\ref{tab:zh_medical}, show that \texttt{BELLE} fine-tuned with {FT} solely on medical data, outperforms the variant trained across three domains in Table~\ref{tab:Chinese_performance}. 
This contrast confirms the presence of inter-domain confusion.

\paragraph{REGA Benefits Both General and Domain Performance.}
However, LLMs fine-tuned using the {REGA} strategy exhibit superior domain-specific performance compared with baselines while maintaining a higher level of general abilities.
To explore why the model trained {REGA} is better, we conduct further analysis in the following sections.

\section{Further Analysis}\label{sec:abalation}
In this section, we analyze the effects of the three components of {REGA}.

\subsection{Effects of Self-Distillation} \label{sec:self-d}
\textbf{Self-Distillation effectively alleviates the catastrophic forgetting of generic abilities.}
As depicted in Table~\ref{tab:Chinese_performance} and Table~\ref{tab:english_performance}, the models trained with strategies with the {Self-Distillation} component (i.e., {FTSD}, {REGA}) achieve higher general scores than those trained with {FT}.
For example, \texttt{Vicuna} achieves a score of 5.68 on the MT-Bench, which notably exceeds the 4.57 of the same model using the {FT}.
This can prove that blending the training corpus with a self-distilled instruction dataset can alleviate the tendency of LLMs to forget their generic capabilities during the training process. 

Furthermore, we also observed a disparity in domain-specific effectiveness when employing the {FTSD} to \texttt{BELLE} and \texttt{Vicuna}.
As illustrated in Table~\ref{tab:Chinese_performance}, the domain-specific performance of both \texttt{BELLE-7B} and \texttt{BELLE-13B}, when trained using the {FTSD} strategy, is inferior to that of models trained under the {FT} approach. 
Conversely, the performance of \texttt{Vicuna} surpasses that of the {FT} configuration. 
We attribute this phenomenon to the English domain data excessively impairing general performance \texttt{Vicuna}, more than the Chinese domain data to \texttt{BELLE}.
This is reflected in poor outcomes in {FQA} where the unigram-F1 score is low.
Besides the limited diversity in English datasets (only 4 compared to 11 Chinese datasets) and the frequent requirement for shorter text responses might also be contributing factors to this issue.

\begin{figure*}[t]
\centering
\includegraphics[width=1\textwidth]{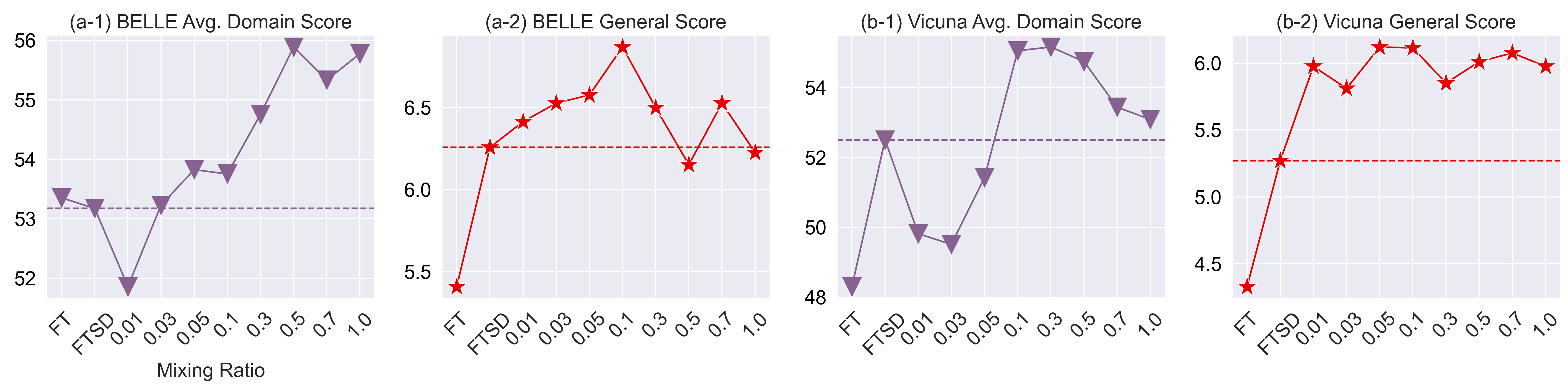}
\caption{General and domain performance of \texttt{BELLE-7B} and \texttt{Vicuna-7B} trained with a varied mixing ratio of {REGA}.}
\label{fig:mix_rate}
\vspace{-1mm}
\end{figure*}

\subsection{Effects of Role Prompting}\label{sec:role-p}

Figure~\ref{fig:rega_role} displays the performance of models trained using {REGA} with different role prompts and compares it to the {FTSD} method.
Figure~\ref{fig:FTRP_COMP} shows the performance of models trained with {FTRP} in response to different role prompts, with comparisons to both {FT} and {FTSD} methods.
These two figures allow us to conclude the following interesting observations:

(1) \textbf{Role Prompts alleviates inter-domain confusion.} 
The {FTRP} strategy outperforms those models trained with {FT} and {FTSD}  across all tested domains.
This superior performance is directly linked to the implementation of domain-specific role prompts throughout training and inference periods.
This proves that Role prompts are crucial for assisting LLMs in recognizing and processing instructions tailored to specific domains by explicitly differentiating them. 

(2) \textbf{Role Prompts elicit abilities within target domains.} 
LLMs, such as \texttt{BELLE}, trained with {REGA} or {FTRP} exhibit higher performance in the medical domain than the other two domains when utilizing the medical role prompt $p_m$, as shown in Figure~\ref{fig:rega_role} and Figure~\ref{fig:FTRP_COMP}.
This is also the same situation for the other two domain-specific role prompts.

\vspace{-2mm}
\subsection{Effects of Role Integration}\label{sec:role-i}

The above section proves that {Role Prompting} can effectively alleviate inter-domain confusion, leading to clear task distinction for the models. 
Concurrently, {Role Integration simplifies the inference process by removing the need for prompt selection and still ensures high performance across various areas.} 
This is evident in Figure~\ref{fig:rega_role}, where top performance typically corresponds with the matrix's diagonal and the REGA$^c$ row.

Moreover, we have the following observations from Figure~\ref{fig:rega_role}:
(1) Adding only 10\% of domain data to the central prompt is sufficient for {REGA}$^c$ to exceed the domain performance of other role prompts, which use the full domain data set. 
(2) Even with access to the entire general and domain datasets, {FTSD} lags behind {REGA}$^c$ across all domains. 
This indicates that the {REGA} model's domain proficiency isn't just a product of shared domain data; there's a clear contribution of knowledge transfer from domain-specific prompts.

Taken together, we argue that the knowledge transfer exists in the {REGA}-tuned model, which flows from the domain role prompts to the central prompt with the help of the shared domain data.

\begin{table}[t]
\centering
\small
\begin{tabular}{c|c|ccc}
\toprule
 \multirow{2}{*}{Model}  & \multicolumn{1}{c}{General} & \multicolumn{3}{c}{Medicine}  \\
\cmidrule(lr){2-2}\cmidrule(lr){3-5} 
&  CGev & QQ & TC & MQA \\ 
\midrule
0-shot  &7.42 & 8.80 & 1.60 & 12.32\\
FT & 5.52 & 84.62 & \textbf{83.60} & \textbf{39.60}\\
FTSD & 6.55 & 82.60 & 79.20 & 36.65\\
REGA$^c$ & \textbf{6.82} &\textbf{{86.30}} & {83.14} & {37.38}\\
\rowcolor{green!20}\multicolumn{5}{c}{\textit{REGA with Different Role Prompts}} \\

REGA$^m$ & 6.33& 84.50 & 81.20 & 39.17\\
\bottomrule
\end{tabular}
\caption{Performance of Belle on medicine and the best scores are in \textbf{bold}. The mixing ratio of REGA is 0.1.}
\label{tab:zh_medical}
\vspace{-2mm}
\end{table}

\section{Discussion}
\subsection{REGA on Single Domain}

We further extend our training to include the \texttt{BELLE-7B} model only within the medical domain, employing the strategies outlined in Section~\ref{sec:baselines}. 
The outcomes of these experiments are detailed in Table~\ref{tab:zh_medical}. 
Analysis of the data presented in Table~\ref{tab:zh_medical} leads us to two key insights: firstly, the inter-domain confusion that can hamper performance is mitigated when focusing on a single domain, as evidenced by the \textbf{FT} approach yielding better results within the medical domain compared to training across multiple domains in Table~\ref{tab:Chinese_performance}. 
Secondly, the {REGA} strategy continues to demonstrate its efficacy by both reducing the loss of general language capabilities and enhancing the model's performance in the domain-specific context.
This indicates that {REGA} still brings significant performance gains even when there is only a single-domain training requirement.

\subsection{Choice of Mixing Ratio}
Then we explore the impact of the mixing ratio $r$ in {Role Integration} (shown in Figure~\ref{fig:mix_rate}). 
We have two observations:
(1) Although a low mixing ratio (such as 0.01) is not enough for {Role Integration} to make {REGA} excel {FT} and {FTSD} in domain test sets, its generic abilities still stay at a superior position compared to other two methods.
(2) The performance of the model train with {REGA} fluctuates with the change of mixing ratio, however, it still surpasses {FT} and {FTSD} by a large margin.
As for the choice of mixing ratio, we recommend a safe interval $[0.05,0.3]$  to simultaneously achieve higher domain and general performance.  

\section{Conclusion}
In this paper, we attempt to strike a balance between domain specialization and generic abilities while adapting LLMs to multiple domains.
Specifically, we propose the {REGA}, which consists of {Self-Distillation} to alleviate the catastrophic forgetting, {Role Prompting} to separate each domain while assigning each role prompt with domain-specific abilities to avoid inter-domain confusion, and {Role Integration} to transfer the domain-specific abilities from the domain-specific role prompt to the central prompt.
Extensive experiments on plenty of datasets and LLMs demonstrate the effectiveness and efficiency of our proposed method.


\section*{Limitations}

In this paper, we introduce the \textbf{REGA} method for studying how to enhance LLMs with capabilities across multiple domains. 
However, \textbf{REGA} relies on pre-existing high-quality instruction sets to build general-domain exemplars. 
The quality of the instruction set determines the retention of the model's general capabilities. 
In this paper, we have made an effort to use open-source, high-quality data, as cited in the previous section.

\section*{Ethics Statement}
In this paper, the datasets and models used are open-source and do not involve any issues related to privacy or contain harmful information. 
The approach proposed aims to enhance the domain capabilities of LLMs, focusing on improving their response accuracy and consistency. 
Additionally, all open-source resources employed in this research are cited or their sources explicitly stated. 
Accordingly, the models we have developed, which are built upon these open-source resources, do not present ethical concerns.

\bibliography{custom}

\clearpage

\appendix

\section{Dataset Statistics} \label{app:data_statis}

In this section, we illustrate the datasets we utilized in Table~\ref{tab:en_data_statis} and Table~\ref{tab:zh_data_statis}, including 5 English datasets and 11 Chinese datasets in three domains, Medicine, Law, and Finance. 
Moreover, the evaluation metrics are also presented in the table.

\section{Prompt Settings}
In this section, we introduce the setting of role prompts in Table~\ref{tab:role_prompts} and the setting of the evaluation prompt in Table~\ref{tab:eval_zh_prompt}  of using GPT-4 to judge the Chinese LLMs' general performance.
\begin{table}[!t]
\small
    \centering
    \colorbox{gray!8}{
    \begin{tabular}{@{}p{7.3cm}}
    \textbf{Medicine:} You are a knowledgeable assistant in the domain of healthcare and medicine, providing detailed answers to medical questions and successfully completing tasks in the medical domain.\\\\
    
    \textbf{Law:} You are a knowledgeable assistant in the domain of law, and you provide detailed answers to users' legal inquiries and other legal requests. You excel at completing tasks in the legal domain.\\\\
    \textbf{Finance:} You are a knowledgeable assistant in the domain of finance, capable of providing detailed answers to users' financial questions and completing tasks in the finance domain very well.\\\\
    \end{tabular}}
    \caption{Role prompts used in \textbf{REGA}. The Chinese version is translated from the above English prompts. The central prompt follows the original setting of LLMs. }
    \label{tab:role_prompts}
\end{table}

\begin{table}[!t]
\small
    \centering
    \colorbox{gray!8}{
    \begin{tabular}{@{}p{7.3cm}}

    请评价AI助手对用户问题的回复质量。\\
    - - -\\
    问题：\{\}\\\\
    - - -\\
    助手：\{\}\\\\
    请分析助手的回复（综合考虑安全性、通顺性、相关性、正确性、信息性、专业性等）。\\
    然后判断每个助手的回复是否存在以下错误：\\
    无意义的重复\\
    语句截断\\
    不当的多语混用\\
    语言不规范\\
    回复与问题不相关\\
    事实错误\\
    违反逻辑规则\\
    未遵循指令或约束\\
    最后给每个助手的回复评分，最高10分，最低0分。\\\\
    请按照以下JSON格式回答，对于错误判断，1代表存在相应错误，0代表不存在：\\
    {"分析": "...", "助手": { "无意义的重复": ?,  "语句截断": ?,    ...    "评分": ?  } }
    
    \end{tabular}}
    \caption{Prompts we used to prompt GPT-4 to evaluate the general performance of LLMs. We request GPT-5 to give a numerical score ranging from 0 to 10.}
    \label{tab:eval_zh_prompt}
\end{table}
\begin{table}[ht]
\centering
\small
\begin{tabular}{ccccc}
\toprule
& \multicolumn{2}{c}{Medical} & \multicolumn{1}{c}{Law} & \multicolumn{1}{c}{Finance} \\

\cmidrule(lr){2-3}\cmidrule(lr){4-4}\cmidrule(lr){5-5}
 & PMQA & MMQA & CHQA & FBQA \\ 
\midrule
\rowcolor[rgb]{0.93,0.93,0.93}\multicolumn{5}{l}{\textit{Training}} \\

Nums. &1,000 & 10,000 & 10,000 &  10,000\\
P. Length & 253.3 & 10.38 &2058.5 &62.6\\
R. Length & 43.2 & 55.95 &1.0 &1034.6\\
\rowcolor[rgb]{0.93,0.93,0.93}\multicolumn{5}{l}{\textit{Testing}} \\

Nums. & 50 & 500 & 500 &  500\\
P. Length & 256.7 &10.25& 1925.7 & 63.0\\
R. Length & 41.1 & 48.65& 1.0 &1034.5\\
\midrule

Metrics  &Acc. & Acc. & Acc. & uF1\\
\bottomrule
\end{tabular}
\caption{Statistics of 5 English datasets. \textbf{P. Length} and \textbf{R. Length} represents the average length of prompts and responses respectively. \textbf{Acc.} means accuracy and the \textbf{uF1} indicates the uni-gram-F1 score.}
\label{tab:en_data_statis}
\end{table}

\begin{table*}[ht]
\centering
\small
\begin{tabular}{c ccc cc cccccc}
\toprule
 & \multicolumn{3}{c}{Medical} & \multicolumn{2}{c}{Law} & \multicolumn{6}{c}{Finance} \\
\cmidrule(lr){2-4}\cmidrule(lr){5-6}\cmidrule(lr){7-12}
  & QQ & TC & MQA & LQA & LS & FNA & FQA & FNL & FRE & FFE &FSP\\ 
\midrule
\rowcolor[rgb]{0.93,0.93,0.93}\multicolumn{12}{l}{\textit{Training}} \\

Nums   & 14,500 &14,110 &28,914 &4,372 &5,235 & 5,000 & 5,000 & 5,000 & 5,000 & 5,000 & 4,000\\
Q. Length & 83.9 &709.3 &31.4 &67.3 &1722.7 & 215.2 & 304.4 &196.4 &282.5 &62.8 &282.8\\
A. Length & 1.0 &12.8 &119.4 &136.0 &247.1 & 25.4 &6.3 & 5.2 & 3.5 &2.0 &7.5\\

\rowcolor[rgb]{0.93,0.93,0.93}\multicolumn{12}{l}{\textit{Testing}} \\
Nums  & 500 & 500  & 1,000 & 500 &500 & 3,600 & 2,469 & 884 & 1,489 &2,020&500\\
Q. Length & 83.6 & 708.6  & 31.5 & 67.0& 1691.5 & 197.9 &301.2 & 189.3 &283.5 &62.8 &300.0\\
A. Length & 1.0 & 12.8  & 122.1 & 137.6&250.7 & 26.0 &6.3 & 5.1  &3.5  &2.0 &6.7  \\

\midrule
Metrics  & Acc. & uF1 & uF1 & uF1 & uF1 & uF1 & uF1 & Acc. & Acc. & Acc. & uF1\\

\bottomrule
\end{tabular}
\caption{Statistics of 11 Chinese datasets. \textbf{P. Length} and \textbf{R. Length} represent the average length of prompts and responses respectively. \textbf{Acc.} means accuracy and the \textbf{uF1} indicates the uni-gram-F1 score. }
\label{tab:zh_data_statis}
\end{table*}


\section{Genenal Performance Evaluation}
The CGev dataset encompasses a range of tasks. The task and instance distribution are as follows:
\textit{Coding}: 19; 
\textit{Information Extraction}: 30;
\textit{Classification}: 31;
\textit{Creative Writing}: 56;
\textit{Recommendation}: 50;
\textit{Dialogues}: 52;
\textit{Knowledge Testing}: 110;
\textit{Context-based Question Answering}: 23;
\textit{Open-domain Question Answering}: 23;
\textit{Rejection}: 25;
\textit{Summarization}: 51;
\textit{Math}: 55;
\textit{Planning}: 17;
\textit{Language Reasoning}: 60;
\textit{Writing}: 47.

\end{CJK*}
\end{document}